\documentclass[conference]{IEEEtran}
\IEEEoverridecommandlockouts

\usepackage{cite}
\usepackage{amsmath,amssymb,amsfonts}
\usepackage{graphicx}
\usepackage{textcomp}
\usepackage{xcolor}
\usepackage{tikz}
\usepackage{balance}
\usetikzlibrary{positioning, arrows.meta, fit, backgrounds}

\usetikzlibrary{
positioning,
arrows.meta,
fit,
backgrounds,
calc,
shapes.geometric,
shadows
}

\definecolor{deepblue}{HTML}{1E3A8A}
\definecolor{softblue}{HTML}{DBEAFE}
\definecolor{softcyan}{HTML}{CFFAFE}
\definecolor{softgreen}{HTML}{DCFCE7}
\definecolor{softorange}{HTML}{FFEDD5}
\definecolor{softpurple}{HTML}{EDE9FE}
\definecolor{softred}{HTML}{FEE2E2}
\definecolor{softgray}{HTML}{F8FAFC}
\definecolor{linegray}{HTML}{334155}

\def\BibTeX{{\rm B\kern-.05em{\sc i\kern-.025em b}\kern-.08em
    T\kern-.1667em\lower.7ex\hbox{E}\kern-.125emX}}

\usepackage{cite}
\usepackage{amsmath,amssymb,amsfonts}
\usepackage{algorithm}
\usepackage{algpseudocode}
\usepackage{booktabs}
\usepackage{array}
\usepackage{url}
\usepackage{xspace}

\providecommand{\MedRLM}{\textsc{MedRLM}\xspace}
\newcommand{\E}{\mathcal{E}}
\newcommand{\G}{\mathcal{G}}
\newcommand{\R}{\mathcal{R}}
\newcommand{\M}{\mathcal{M}}
\newcommand{\A}{\mathcal{A}}
\newcommand{\D}{\mathcal{D}}

\usepackage[T1]{fontenc}
\usepackage{lmodern}
\usepackage{xcolor}
\usepackage{tikz}
\usetikzlibrary{
arrows.meta,
positioning,
calc,
fit,
backgrounds,
shapes.geometric,
decorations.pathreplacing,
shadows
}

\definecolor{mNavy}{HTML}{0B1F3A}
\definecolor{mBlue}{HTML}{1F5EFF}
\definecolor{mCyan}{HTML}{00A6D6}
\definecolor{mTeal}{HTML}{00A389}
\definecolor{mGreen}{HTML}{2EAD67}
\definecolor{mAmber}{HTML}{F5A623}
\definecolor{mOrange}{HTML}{F97316}
\definecolor{mRed}{HTML}{E63946}
\definecolor{mPurple}{HTML}{7C3AED}
\definecolor{mSlate}{HTML}{475569}
\definecolor{mLight}{HTML}{F8FAFC}
\definecolor{mPaper}{HTML}{FFFFFF}

\tikzset{
  >=Latex,
  medTiny/.style={
    font=\fontsize{5.8}{6.6}\selectfont,
    align=center
  },
  medSmall/.style={
    font=\fontsize{6.7}{7.5}\selectfont,
    align=center
  },
  medLabel/.style={
    font=\fontsize{7.3}{8.2}\selectfont\bfseries,
    align=center
  },
  medTitle/.style={
    font=\fontsize{10.5}{11.5}\selectfont\bfseries,
    text=mNavy
  },
  medStage/.style={
    font=\fontsize{6.0}{6.8}\selectfont\bfseries,
    text=white,
    rounded corners=4pt,
    inner xsep=5pt,
    inner ysep=2pt
  },
  medBoard/.style={
    rounded corners=12pt,
    draw=mSlate!22,
    line width=.55pt,
    fill=mLight,
    drop shadow={
      opacity=.12,
      shadow xshift=.8pt,
      shadow yshift=-.8pt
    }
  },
  medBox/.style={
    rounded corners=6pt,
    line width=.62pt,
    inner sep=3.3pt,
    align=center,
    drop shadow={
      opacity=.10,
      shadow xshift=.6pt,
      shadow yshift=-.6pt
    }
  },
  medInput/.style={
    medBox,
    draw=mBlue!60!black,
    fill=mBlue!5,
    text width=2.55cm,
    minimum width=2.78cm,
    minimum height=.68cm,
    font=\fontsize{6.1}{6.7}\selectfont
  },
  medAgent/.style={
    medBox,
    draw=mTeal!62!black,
    fill=mTeal!6,
    text width=1.78cm,
    minimum width=2.02cm,
    minimum height=.72cm,
    font=\fontsize{6.1}{6.9}\selectfont
  },
  medCore/.style={
    medBox,
    draw=mPurple!70!black,
    fill=mPurple!7,
    text width=3.65cm,
    minimum width=3.95cm,
    minimum height=.98cm,
    font=\fontsize{6.5}{7.3}\selectfont
  },
  medGraph/.style={
    medBox,
    draw=mAmber!80!black,
    fill=mAmber!8,
    text width=3.45cm,
    minimum width=3.78cm,
    minimum height=4.10cm,
    font=\fontsize{6.0}{6.8}\selectfont
  },
  medOut/.style={
    medBox,
    draw=mGreen!65!black,
    fill=mGreen!7,
    text width=2.95cm,
    minimum width=3.20cm,
    minimum height=.78cm,
    font=\fontsize{6.2}{7.0}\selectfont
  },
  medSafety/.style={
    medBox,
    draw=mRed!72!black,
    fill=mRed!6,
    text width=3.05cm,
    minimum width=3.25cm,
    minimum height=.78cm,
    font=\fontsize{6.1}{6.9}\selectfont
  },
  medArrow/.style={
    -{Latex[length=2.2mm]},
    line width=.62pt,
    draw=mNavy!72
  },
  medSoftArrow/.style={
    -{Latex[length=2.0mm]},
    line width=.52pt,
    draw=mSlate!62
  },
  medRecurArrow/.style={
    -{Latex[length=2.2mm]},
    line width=.68pt,
    draw=mPurple!82,
    dashed
  },
  medEvidenceArrow/.style={
    -{Latex[length=2.1mm]},
    line width=.62pt,
    draw=mAmber!82!black
  },
  medRiskArrow/.style={
    -{Latex[length=2.1mm]},
    line width=.64pt,
    draw=mGreen!72!black
  },
  medNode/.style={
    circle,
    draw=mNavy!35,
    line width=.42pt,
    minimum size=.22cm,
    inner sep=0pt,
    fill=white
  }
}

\usepackage{cite}
\usepackage{amsmath,amssymb,amsfonts}
\usepackage{graphicx}
\usepackage{textcomp}
\usepackage{xcolor}
\usepackage{booktabs}
\usepackage{array}
\usepackage{url}
\usepackage{xspace}
\usepackage{tikz}
\usetikzlibrary{positioning,arrows.meta,calc}

\definecolor{mBlue}{HTML}{1F5EFF}
\definecolor{mCyan}{HTML}{00A6D6}
\definecolor{mTeal}{HTML}{00A389}
\definecolor{mAmber}{HTML}{F5A623}
\definecolor{mOrange}{HTML}{F97316}
\definecolor{mPurple}{HTML}{7C3AED}
\definecolor{mSlate}{HTML}{475569}
\definecolor{mLight}{HTML}{F8FAFC}

\title{MedRLM: Recursive Multimodal Health Intelligence for Long-Context Clinical Reasoning, Sensor-Guided Screening, Evidence-Grounded Decision Support, and Community-to-Tertiary Referral Optimization}

\author{
\IEEEauthorblockN{
Aueaphum Aueawatthanaphisut\IEEEauthorrefmark{1}
}
\IEEEauthorblockA{
School of Information, Computer, and Communication Technology\\
Sirindhorn International Institute of Technology, Thammasat University\\
Pathum Thani, Thailand\\
\IEEEauthorrefmark{1}aueawatth.aue@gmail.com,
}
}

\begin{document}
\maketitle

\begin{abstract}
Real-world clinical decision support requires reasoning over heterogeneous and longitudinal patient information rather than answering isolated medical questions. However, current medical large language models and retrieval-augmented generation systems often rely on single-step prompting or retrieval, which can be fragile when clinical evidence is distributed across long electronic health records, medical images, sensor streams, guidelines, and referral constraints. This paper proposes \MedRLM, a Recursive Multimodal Health Intelligence framework for long-context clinical reasoning, sensor-guided screening, and community-to-tertiary referral support. Instead of compressing all patient information into one prompt, \MedRLM treats the patient case as an external clinical environment that can be recursively inspected, decomposed, retrieved, verified, and synthesized. The framework coordinates specialized agents for clinical text, longitudinal EHR, medical imaging, physiological sensor signals, guideline retrieval, uncertainty auditing, and referral planning. It further introduces a Clinical Evidence Graph Memory to connect patient-specific observations with retrieved evidence, standardized definitions, sensor-derived biomarkers, and referral criteria. A sensor-guided recursive triggering mechanism activates deeper reasoning when abnormal physiological or behavioral patterns are detected, while uncertainty-gated refinement supports clinician review for high-risk or low-confidence cases. We also outline a real-data evaluation design using public and credentialed clinical datasets spanning EHR, radiology, ECG, ICU time series, and referral-proxy outcomes. \MedRLM aims to move medical AI from static question answering toward auditable, multimodal, and workflow-aware clinical decision support.
\end{abstract}


\section{Introduction}
\label{sec:introduction}

Artificial intelligence is increasingly being explored as a decision-support layer for healthcare systems, particularly in clinical question answering, medical image interpretation, patient triage, and evidence-based recommendation. Recent medical large language models (LLMs) and biomedical foundation models have demonstrated strong capability in medical knowledge encoding, expert-level medical question answering, and multimodal biomedical reasoning~\cite{b8,b9,b10,b11}. However, real-world clinical decision support is rarely a single-turn question-answering task. A patient case often contains long and heterogeneous information, including clinical notes, longitudinal electronic health records (EHRs), medical images, wearable or embedded sensor signals, guideline documents, and referral constraints. Compressing these data into a single prompt can lead to context loss, hallucination, weak traceability, and unreliable clinical reasoning.

A key limitation of current long-context LLMs is that larger context windows do not necessarily imply reliable use of all relevant evidence. Prior studies show that language models can suffer from long-context degradation, where information hidden in the middle of long inputs is underutilized, and benchmark performance declines as context length and task complexity increase~\cite{b4,b5}. Recursive Language Models (RLMs) address this issue by treating long prompts as an external environment that can be inspected, decomposed, and processed through recursive model calls~\cite{b1}. Follow-up studies further suggest that recursion should be structured and uncertainty-aware to avoid inefficient or semantically weak reasoning trajectories~\cite{b2,b3}. These advances provide a promising foundation for clinical AI, where patient information is naturally distributed across multiple sources and often requires iterative evidence gathering.

In parallel, retrieval-augmented generation (RAG) has become a practical approach for grounding LLM outputs in external knowledge~\cite{b6}. In medicine, RAG is particularly important because clinical recommendations must be factual, traceable, and aligned with up-to-date medical evidence. Medical RAG benchmarks and graph-based medical RAG systems have shown that retrieval can improve medical question answering and reduce unsupported generation~\cite{b12,b14}. Multimodal medical RAG has also improved factuality in medical vision-language models by integrating external medical references with image understanding~\cite{b13}. Nevertheless, most existing medical RAG systems remain retrieval-centric: they retrieve evidence for a query but do not fully model the clinical workflow as a recursive process that integrates long patient history, sensor-derived biomarkers, multimodal evidence, uncertainty estimation, and referral decision-making.

This paper proposes \MedRLM, a Recursive Multimodal Health Intelligence framework for long-context clinical reasoning, sensor-guided screening, and community-to-tertiary referral support. Unlike conventional medical LLM or RAG pipelines, \MedRLM treats heterogeneous patient data as an external clinical environment rather than a monolithic prompt. The system recursively decomposes a clinical query into modality-specific subtasks, invokes specialized agents for text, EHR, image, sensor, guideline, and referral reasoning, constructs an auditable Clinical Evidence Graph Memory, and synthesizes a risk-aware referral recommendation. This design is particularly suited for resource-limited healthcare settings, where community hospitals may require early screening support and structured referral guidance before escalating cases to tertiary care.

The main contributions of this paper are as follows:
\begin{itemize}
    \item We introduce \MedRLM, a recursive multimodal clinical intelligence framework that extends RLM-style long-context inference to healthcare decision support.
    \item We propose a Clinical Evidence Graph Memory that connects patient-specific observations, sensor-derived biomarkers, medical evidence, clinical definitions, and referral criteria into an auditable reasoning structure.
    \item We introduce sensor-guided recursive triggering, where abnormal physiological or behavioral signals initiate deeper reasoning over symptoms, longitudinal records, images, guidelines, and referral rules.
    \item We formulate uncertainty-gated recursive refinement and referral-utility optimization to support safe escalation from community care to specialist or tertiary-care pathways.
\end{itemize}

By combining recursive long-context inference, multimodal clinical reasoning, evidence-grounded retrieval, sensor-based screening, and referral optimization, \MedRLM aims to move medical AI beyond static question answering toward auditable, workflow-aware, and deployment-oriented clinical decision support.

\section{Related Work}
\label{sec:related_work}

\subsection{Long-Context Reasoning and Recursive Language Models}

Long-context reasoning has become a central challenge for LLM-based systems. Although modern LLMs can accept increasingly long inputs, prior work shows that they may fail to robustly use relevant information when it appears in the middle of a long context~\cite{b4}. LongBench further formalizes long-context evaluation across document question answering, summarization, few-shot learning, synthetic retrieval, and code tasks, showing that long-context understanding remains difficult even for strong models~\cite{b5}. These findings suggest that simply increasing the context window is insufficient for reliable reasoning over large and heterogeneous clinical data.

Recursive Language Models provide an inference-time alternative by externalizing long prompts and allowing the model to programmatically examine, decompose, and recursively process smaller pieces of information~\cite{b1}. SRLM improves this direction by incorporating uncertainty-aware self-reflective program search, while $\lambda$-RLM argues for more structured recursive control using typed functional mechanisms~\cite{b2,b3}. \MedRLM builds on these ideas but shifts the target domain from generic long-context tasks to clinical reasoning. Instead of recursively processing text alone, \MedRLM recursively coordinates clinical notes, EHR timelines, medical images, sensor streams, guidelines, and referral rules.

\subsection{Medical Large Language Models and Biomedical Foundation Models}

Medical LLMs have shown strong potential for clinical knowledge encoding and medical question answering. Med-PaLM demonstrated that LLMs can be aligned toward medical question answering using domain-specific prompting and evaluation~\cite{b8}, while Med-PaLM 2 further improved expert-level medical QA performance through stronger reasoning and grounding strategies~\cite{b9}. Biomedical vision-language systems such as LLaVA-Med extend instruction-following capabilities to biomedical images~\cite{b10}. Generalist biomedical AI systems such as Med-PaLM Multimodal further demonstrate the potential of unified models that handle clinical language, imaging, genomics, and other biomedical modalities~\cite{b11}.

Despite this progress, most medical LLMs are still primarily evaluated as answer-generation systems rather than workflow-level clinical reasoning systems. They often lack explicit mechanisms for long patient histories, sensor-triggered reasoning, referral pathway optimization, and auditable evidence synthesis. \MedRLM addresses this gap by positioning the LLM as a recursive clinical controller that coordinates specialized agents and external clinical memory rather than acting as a single monolithic answer generator.

\subsection{Retrieval-Augmented Generation and Medical Evidence Grounding}

RAG combines parametric language models with external non-parametric memory, improving factuality, updateability, and source traceability~\cite{b6}. ReAct further shows that interleaving reasoning and action can help language models retrieve information, update plans, and interact with external environments~\cite{b7}. In medical settings, retrieval is essential because clinical decisions require evidence-based grounding rather than plausible but unsupported generation. MEDRAG and MIRAGE provide systematic evaluation of medical RAG systems and demonstrate the importance of selecting appropriate corpora, retrievers, and backbone models for medical QA~\cite{b12}.

Recent medical graph-based RAG methods improve evidence grounding by linking user documents to credible medical sources and definitions~\cite{b14}. Multimodal medical RAG further addresses factual hallucination in medical vision-language models by using domain-aware retrieval and adaptive retrieved-context selection~\cite{b13}. However, these systems largely focus on improving answer factuality for medical QA or vision-language tasks. \MedRLM extends this line of work by integrating retrieval into a recursive clinical workflow, where retrieved evidence is not only used to answer a question but also to update risk estimation, uncertainty assessment, sensor interpretation, and referral planning.

\subsection{Longitudinal EHR, Multimodal Patient Data, and Referral Support}

Longitudinal EHR modeling is critical for realistic clinical AI because patient risk often emerges from temporal patterns rather than isolated observations. EHRSHOT highlights the need for benchmarks and foundation models that evaluate few-shot clinical prediction over longitudinal structured EHR data~\cite{b15}. However, EHR-only modeling does not fully address real-world community screening scenarios, where decisions may depend on symptoms, low-cost sensors, smartphone images, clinical guidelines, and local referral capacity.

Existing multimodal biomedical AI systems demonstrate the value of integrating multiple medical data types~\cite{b10,b11}, but few frameworks explicitly connect multimodal reasoning with community-to-tertiary referral decisions. \MedRLM fills this gap by combining longitudinal patient representation, sensor-derived digital biomarkers, multimodal evidence retrieval, and referral utility optimization within a single recursive reasoning framework. This makes the proposed framework distinct from prior medical LLM, RAG, and multimodal AI systems: its core objective is not only to answer clinical questions, but to produce auditable, risk-aware, and context-sensitive referral support.

\section{Methodology}
\label{sec:methodology}

This section presents \MedRLM, a recursive multimodal health intelligence framework for long-context clinical reasoning, sensor-guided screening, and community-to-tertiary referral support. The main methodological principle is that heterogeneous patient data should not be compressed into a single long prompt. Instead, patient records, clinical notes, images, sensor streams, guidelines, and referral protocols are treated as an external clinical environment that can be recursively inspected, decomposed, retrieved, verified, and synthesized. This design is motivated by recursive language modeling, uncertainty-aware recursive search, structured recursive control, retrieval-augmented generation, reasoning--acting agents, medical large language models, multimodal biomedical foundation models, medical graph retrieval, multimodal medical RAG, and longitudinal EHR benchmarks~\cite{b1,b2,b3,b4,b5,b6,b7,b8,b9,b10,b11,b12,b13,b14,b15}.

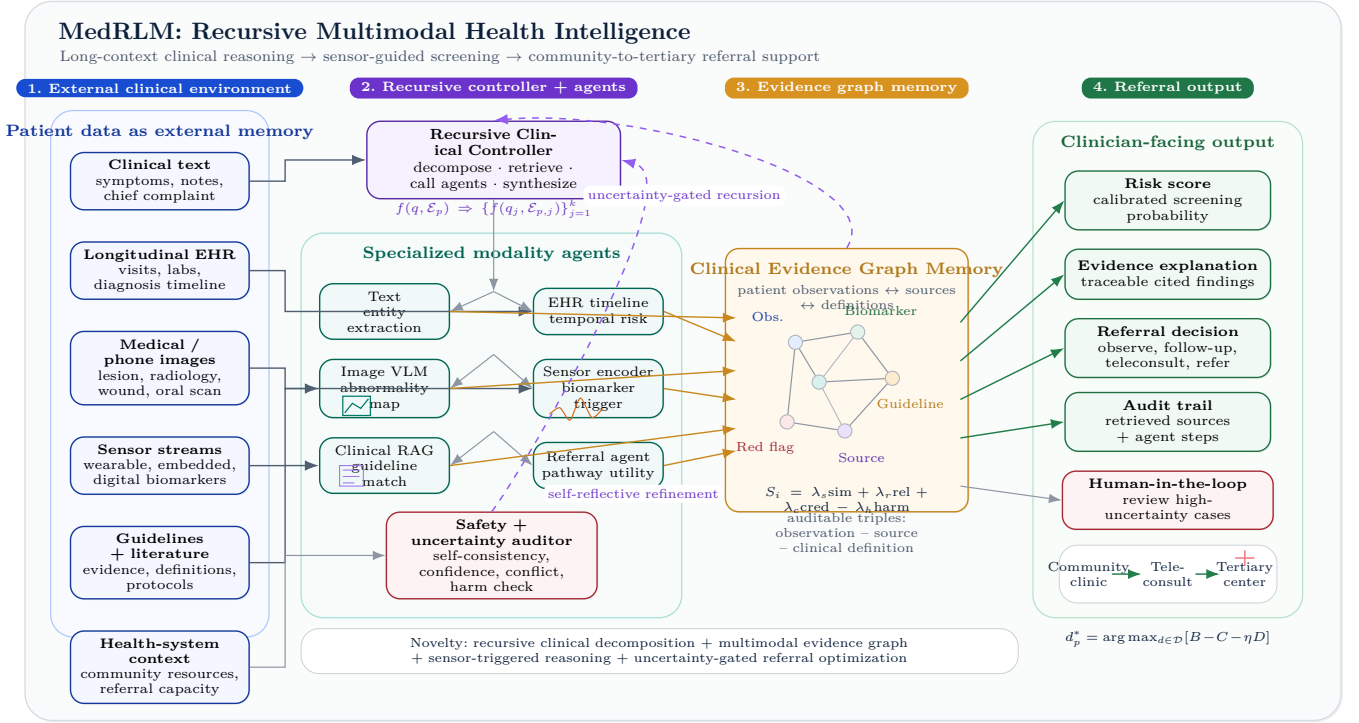
\begin{figure*}[t]
\centering
\resizebox{0.98\textwidth}{!}{%
\begin{tikzpicture}[x=1cm,y=1cm]

\begin{scope}[on background layer]
  \draw[medBoard] (-.45,-.55) rectangle (20.05,10.65);
  \fill[rounded corners=12pt, mNavy!2]
  (-.35,-.45) rectangle (19.95,10.55);
\end{scope}

\node[medTitle, anchor=west] at (-.05,10.16)
{MedRLM: Recursive Multimodal Health Intelligence};

\node[medSmall, text=mSlate, anchor=west] at (-.03,9.78)
{Long-context clinical reasoning $\rightarrow$ sensor-guided screening $\rightarrow$ community-to-tertiary referral support};

\node[medStage, fill=mBlue!82!black] at (1.65,9.30)
{1. External clinical environment};

\node[medStage, fill=mPurple!86!black] at (6.85,9.30)
{2. Recursive controller + agents};

\node[medStage, fill=mAmber!88!black] at (12.35,9.30)
{3. Evidence graph memory};

\node[medStage, fill=mGreen!68!black] at (17.35,9.30)
{4. Referral output};

\draw[rounded corners=10pt, draw=mBlue!35, fill=mBlue!3, line width=.55pt]
(-.05,0.75) rectangle (3.35,8.95);

\node[medLabel, text=mBlue!60!black] at (1.65,8.62)
{Patient data as external memory};

\node[medInput] (text) at (1.65,7.85)
{\textbf{Clinical text}\\ symptoms, notes, chief complaint};

\node[medInput, below=4.8mm of text] (ehr)
{\textbf{Longitudinal EHR}\\ visits, labs, diagnosis timeline};

\node[medInput, below=4.8mm of ehr] (img)
{\textbf{Medical / phone images}\\ lesion, radiology, wound, oral scan};

\node[medInput, below=4.8mm of img] (sens)
{\textbf{Sensor streams}\\ wearable, embedded, digital biomarkers};

\node[medInput, below=4.8mm of sens] (guides)
{\textbf{Guidelines + literature}\\ evidence, definitions, protocols};

\node[medInput, below=4.8mm of guides] (constraints)
{\textbf{Health-system context}\\ community resources, referral capacity};





\node[medCore] (controller) at (6.85,8.18)
{\textbf{Recursive Clinical Controller}\\
decompose $\cdot$ retrieve $\cdot$ call agents $\cdot$ synthesize};

\node[medTiny, text=mPurple!82!black, text width=3.7cm] at (6.85,7.42)
{$f(q,\mathcal{E}_p)\Rightarrow \{f(q_j,\mathcal{E}_{p,j})\}_{j=1}^{k}$};

\draw[rounded corners=10pt, draw=mTeal!35, fill=mTeal!3, line width=.55pt]
(3.85,1.05) rectangle (9.78,7.04);

\node[medLabel, text=mTeal!60!black] at (6.82,6.72)
{Specialized modality agents};

\node[medAgent] (a_text) at (5.15,5.82)
{Text\\ entity extraction};

\node[medAgent] (a_ehr) at (8.48,5.82)
{EHR timeline\\ temporal risk};

\node[medAgent] (a_img) at (5.15,4.62)
{Image VLM\\ abnormality map};

\node[medAgent] (a_sens) at (8.48,4.62)
{Sensor encoder\\ biomarker trigger};

\node[medAgent] (a_guides) at (5.15,3.42)
{Clinical RAG\\ guideline match};

\node[medAgent] (a_ref) at (8.48,3.42)
{Referral agent\\ pathway utility};

\node[medSafety] (auditor) at (6.82,2.02)
{\textbf{Safety + uncertainty auditor}\\
self-consistency, confidence, conflict, harm check};

\draw[mTeal!75!black, line width=.55pt]
(4.50,4.20) rectangle (4.93,4.50);

\draw[mTeal!75!black, line width=.55pt]
(4.53,4.24) -- (4.65,4.39) -- (4.78,4.31) -- (4.90,4.46);

\draw[mOrange!85!black, line width=.50pt, smooth]
(7.72,4.22) plot coordinates
{(7.72,4.22) (7.88,4.32) (8.04,4.13) (8.20,4.47) (8.38,4.17) (8.55,4.31)};

\draw[mPurple!70, line width=.48pt]
(4.45,3.10)--(4.82,3.10)--(4.82,3.42)--(4.45,3.42)--cycle;

\draw[mPurple!70, line width=.48pt] (4.50,3.35)--(4.76,3.35);
\draw[mPurple!70, line width=.48pt] (4.50,3.25)--(4.76,3.25);
\draw[mPurple!70, line width=.48pt] (4.50,3.15)--(4.68,3.15);

\node[medGraph] (graphpanel) at (12.35,4.75) {};

\node[medLabel, text=mAmber!72!black] at (12.35,6.47)
{Clinical Evidence Graph Memory};

\node[medTiny, text=mSlate, text width=3.40cm] at (12.35,6.04)
{patient observations $\leftrightarrow$ sources\\
$\leftrightarrow$ definitions};

\node[medNode, fill=mBlue!12]   (g1) at (11.55,5.34) {};
\node[medNode, fill=mGreen!13]  (g2) at (12.52,5.50) {};
\node[medNode, fill=mAmber!22]  (g3) at (13.06,4.78) {};
\node[medNode, fill=mPurple!15] (g4) at (12.32,3.96) {};
\node[medNode, fill=mRed!12]    (g5) at (11.42,4.10) {};
\node[medNode, fill=mTeal!15]   (g6) at (11.92,4.72) {};

\draw[mSlate!62, line width=.52pt]
(g1)--(g2)--(g3)--(g4)--(g5)--(g1)--(g6)--(g3);

\draw[mSlate!45, line width=.42pt]
(g2)--(g6)--(g4);

\node[medTiny, text=mBlue!65!black] at (11.12,5.74)
{Obs.};

\node[medTiny, text=mGreen!60!black] at (12.88,5.84)
{Biomarker};

\node[medTiny, text=mAmber!72!black] at (13.34,4.38)
{Guideline};

\node[medTiny, text=mPurple!75!black] at (12.58,3.54)
{Source};

\node[medTiny, text=mRed!70!black] at (11.08,3.70)
{Red flag};

\node[medTiny, text=mNavy, text width=3.45cm] at (12.35,2.88)
{$S_i=\lambda_s\mathrm{sim}+\lambda_r\mathrm{rel}
+\lambda_c\mathrm{cred}-\lambda_h\mathrm{harm}$};

\node[medTiny, text=mSlate, text width=3.45cm] at (12.35,2.38)
{auditable triples:\\ observation -- source -- clinical definition};

\draw[rounded corners=10pt, draw=mGreen!35, fill=mGreen!3, line width=.55pt]
(15.25,1.05) rectangle (19.45,8.78);

\node[medLabel, text=mGreen!58!black] at (17.35,8.45)
{Clinician-facing output};

\node[medOut] (risk) at (17.35,7.55)
{\textbf{Risk score}\\ calibrated screening probability};

\node[medOut] (explain) at (17.35,6.40)
{\textbf{Evidence explanation}\\ traceable cited findings};

\node[medOut] (refer) at (17.35,5.25)
{\textbf{Referral decision}\\ observe, follow-up, teleconsult, refer};

\node[medOut] (audit) at (17.35,4.10)
{\textbf{Audit trail}\\ retrieved sources + agent steps};

\node[medSafety] (human) at (17.35,2.88)
{\textbf{Human-in-the-loop}\\ review high-uncertainty cases};

\draw[rounded corners=8pt, draw=mNavy!20, fill=white, line width=.45pt]
(15.66,1.28) rectangle (19.05,2.18);

\node[medTiny, text=mNavy] at (16.12,1.73)
{Community\\clinic};

\node[medTiny, text=mNavy] at (17.35,1.73)
{Tele-\\consult};

\node[medTiny, text=mNavy] at (18.55,1.73)
{Tertiary\\center};

\draw[medRiskArrow] (16.48,1.73)--(16.94,1.73);
\draw[medRiskArrow] (17.78,1.73)--(18.16,1.73);

\draw[mRed!70, line width=.65pt] (18.40,1.98)--(18.70,1.98);
\draw[mRed!70, line width=.65pt] (18.55,1.86)--(18.55,2.10);

\draw[medArrow] (text.east) -- ++(.55,0) |- (controller.west);
\draw[medArrow] (ehr.east) -- ++(.55,0) |- (a_ehr.west);
\draw[medArrow] (img.east) -- ++(.55,0) |- (a_img.west);
\draw[medArrow] (sens.east) -- ++(.55,0) |- (a_sens.west);
\draw[medArrow] (guides.east) -- ++(.55,0) |- (a_guides.west);
\draw[medSoftArrow] (constraints.east) -- ++(.55,0) |- (auditor.west);

\draw[medSoftArrow] (controller.south) -- (6.85,7.14) -- (6.85,6.13);
\draw[medSoftArrow] (6.85,6.13) -- (a_text.east);
\draw[medSoftArrow] (6.85,6.13) -- (a_ehr.west);
\draw[medSoftArrow] (6.85,5.15) -- (a_img.east);
\draw[medSoftArrow] (6.85,5.15) -- (a_sens.west);
\draw[medSoftArrow] (6.85,3.95) -- (a_guides.east);
\draw[medSoftArrow] (6.85,3.95) -- (a_ref.west);

\draw[medEvidenceArrow] (a_text.east) -- (10.63,5.72);
\draw[medEvidenceArrow] (a_ehr.east) -- (10.63,5.35);
\draw[medEvidenceArrow] (a_img.east) -- (10.63,4.90);
\draw[medEvidenceArrow] (a_sens.east) -- (10.63,4.45);
\draw[medEvidenceArrow] (a_guides.east) -- (10.63,4.00);
\draw[medEvidenceArrow] (a_ref.east) -- (10.63,3.65);

\draw[medRiskArrow] (14.12,5.65) -- (risk.west);
\draw[medRiskArrow] (14.12,5.05) -- (explain.west);
\draw[medRiskArrow] (14.12,4.45) -- (refer.west);
\draw[medRiskArrow] (14.12,3.85) -- (audit.west);
\draw[medSoftArrow] (14.12,3.10) -- (human.west);

\draw[medRecurArrow]
(auditor.north) .. controls (7.10,3.02) and (10.02,7.86) .. (controller.east);

\draw[medRecurArrow]
(graphpanel.north) .. controls (13.05,7.90) and (9.00,9.00) .. (controller.north);

\node[medTiny, text=mPurple!85!black, fill=white,
rounded corners=3pt, inner sep=2pt] at (9.82,7.62)
{uncertainty-gated recursion};

\node[medTiny, text=mPurple!85!black, fill=white,
rounded corners=3pt, inner sep=2pt] at (9.03,2.98)
{self-reflective refinement};

\node[medTiny, text=mNavy, text width=3.2cm] at (17.35,.72)
{$d_p^{*}=\arg\max_{d\in\mathcal{D}}[B-C-\eta D]$};

\draw[rounded corners=8pt, draw=mNavy!20, fill=white, line width=.45pt]
(3.85,.18) rectangle (15.02,.88);

\node[medTiny, text=mNavy, text width=10.8cm, align=center] at (9.43,.53)
{Novelty: recursive clinical decomposition + multimodal evidence graph + sensor-triggered reasoning + uncertainty-gated referral optimization};

\end{tikzpicture}%
}
\caption{Architecture of the proposed MedRLM framework. The system treats heterogeneous patient data as an external clinical environment, recursively decomposes clinical queries into modality-specific subtasks, constructs an auditable evidence graph, and produces risk-aware community-to-tertiary referral recommendations.}
\label{fig:medrlm_architecture}
\end{figure*}

\begin{figure*}
    \centering
    \includegraphics[width=1\linewidth]{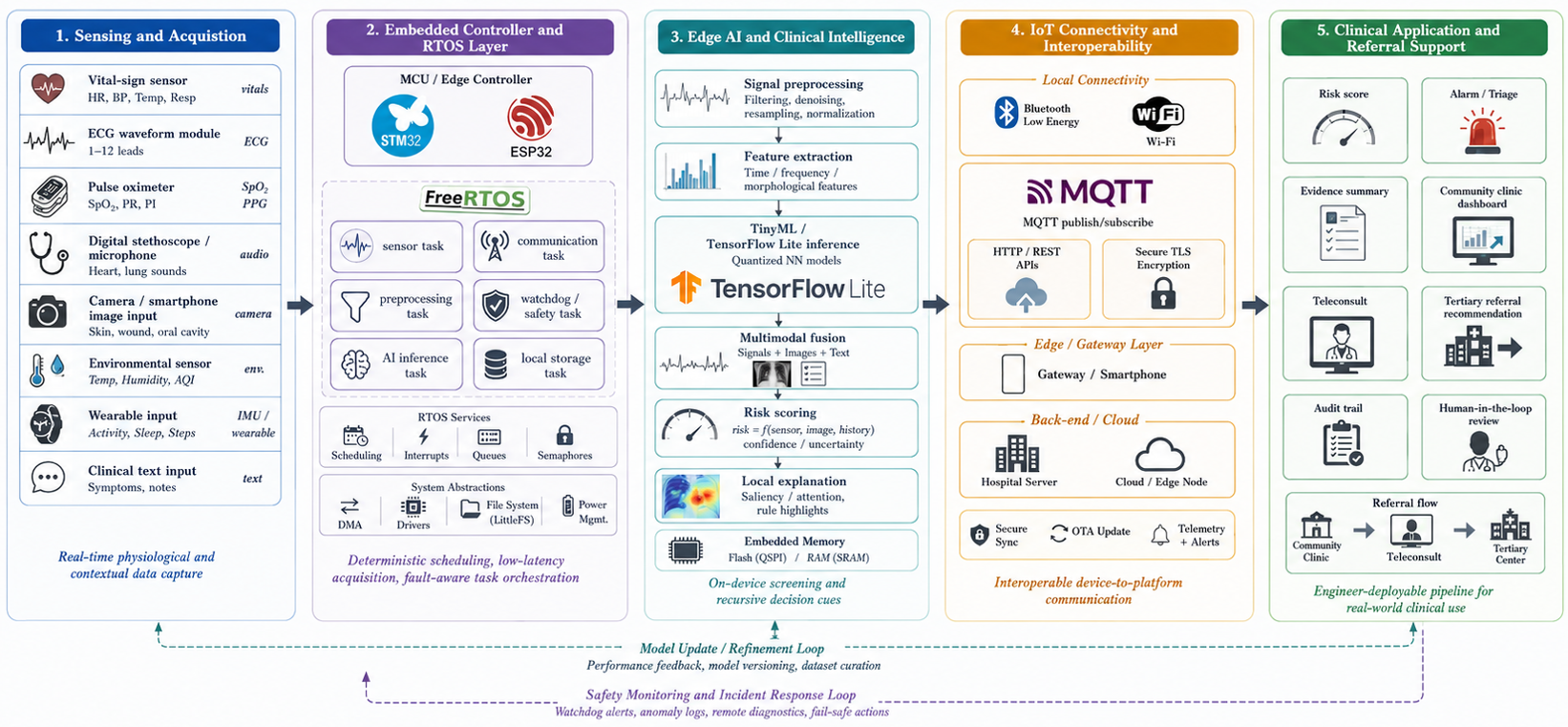}
    \caption{Embedded MedRLM architecture for RTOS-enabled edge AI and IoT-based clinical support. The proposed implementation pipeline begins with real-time sensing and acquisition from physiological, imaging, environmental, wearable, and clinical-text inputs. These data streams are managed by an embedded controller through RTOS-based task orchestration, including sensing, preprocessing, AI inference, communication, watchdog safety, and local storage tasks. The edge intelligence layer performs signal preprocessing, feature extraction, TinyML inference, multimodal fusion, risk scoring, and local explanation before transmitting structured outputs through secure IoT protocols such as BLE, Wi-Fi, MQTT, HTTP/REST, and TLS. The clinical application layer converts device-level intelligence into risk scores, triage alerts, evidence summaries, audit trails, teleconsultation support, and community-to-tertiary referral recommendations.}
    \label{fig:embedded_medrlm_architecture}
\end{figure*}

Fig.~\ref{fig:embedded_medrlm_architecture} illustrates the engineering-level deployment architecture of MedRLM, where heterogeneous physiological and contextual data are acquired by embedded sensors, scheduled through an RTOS layer, processed by on-device TinyML inference, transmitted through secure IoT interfaces, and converted into auditable clinical decision and referral-support outputs.

\subsection{Clinical Environment Formulation}

For each patient case $p$, MedRLM constructs an external clinical environment rather than a single input sequence. The environment is defined as
\begin{equation}
\begin{aligned}
\E_p = \{&X^{\mathrm{text}}_p, X^{\mathrm{ehr}}_p,
X^{\mathrm{img}}_p, X^{\mathrm{sens}}_p, \\
&\G, \R, \mathcal{H}\},
\end{aligned}
\label{eq:environment}
\end{equation}
where $X^{\mathrm{text}}_p$ denotes free-text symptoms, clinical notes, and patient narratives; $X^{\mathrm{ehr}}_p$ denotes longitudinal structured electronic health record events; $X^{\mathrm{img}}_p$ denotes medical or smartphone-acquired images; $X^{\mathrm{sens}}_p$ denotes wearable or embedded sensor streams; $\G$ denotes clinical guideline knowledge; $\R$ denotes referral rules; and $\mathcal{H}$ denotes local healthcare-system constraints, such as community-hospital capability, specialist availability, and tertiary-care accessibility.

Given a clinical query $q$, the objective of MedRLM is to produce a clinically useful output
\begin{equation}
Y_p = (\hat{y}_p, r_p, d_p, \A_p, U_p),
\label{eq:output}
\end{equation}
where $\hat{y}_p$ is the predicted clinical interpretation, $r_p$ is the estimated risk score, $d_p$ is the recommended referral or care-pathway decision, $\A_p$ is the evidence audit trail, and $U_p$ is the uncertainty and safety score. The system is designed as a clinician-support tool, not as an autonomous diagnostic authority.

\subsection{Recursive Multimodal Clinical Controller}

The core of MedRLM is a recursive clinical controller that determines whether a clinical task can be answered directly from retrieved evidence or should be decomposed into smaller subtasks. Let $\kappa(q,\E_p)$ be a context-complexity function that measures the length, heterogeneity, evidence dispersion, and clinical risk level of the case. The recursive controller is defined as
\begin{equation}
\begin{aligned}
\Phi(q,\E_p) =
\begin{cases}
M\big(q,\rho(q,\E_p)\big),
& \kappa(q,\E_p) \leq K, \\
\Omega\!\left(
\left\{\Phi(q_j,\E_{p,j})\right\}_{j=1}^{k}
\right),
& \kappa(q,\E_p) > K,
\end{cases}
\end{aligned}
\label{eq:recursive_controller}
\end{equation}
where $M$ is the base medical language or vision-language model, $\rho(\cdot)$ is the evidence retrieval function, $K$ is the maximum safe context threshold, $q_j$ is a decomposed clinical subquery, $\E_{p,j}$ is the corresponding sub-environment, $k$ is the number of recursive branches, and $\Omega(\cdot)$ is the synthesis operator. This recursion allows MedRLM to process long patient histories and distributed clinical evidence without relying on lossy one-shot summarization.

The complexity score is computed as
\begin{equation}
\begin{aligned}
\kappa(q,\E_p) =
&\gamma_1 L(q,\E_p)
+ \gamma_2 V(\E_p)
+ \gamma_3 D(\E_p) \\
&+ \gamma_4 R_{\mathrm{risk}}(p)
+ \gamma_5 C_{\mathrm{conflict}}(\E_p),
\end{aligned}
\label{eq:complexity}
\end{equation}
where $L(\cdot)$ measures context length, $V(\cdot)$ measures modality diversity, $D(\cdot)$ measures evidence dispersion across the environment, $R_{\mathrm{risk}}(p)$ represents preliminary clinical risk, and $C_{\mathrm{conflict}}(\cdot)$ measures contradiction among available evidence.

\subsection{Recursive Clinical Task Decomposition}

A complex clinical query is decomposed into modality-aware and decision-aware subtasks:
\begin{equation}
\begin{aligned}
\mathcal{Q}_p =
\mathrm{Dec}(q,\E_p) =
\{&q^{\mathrm{text}}, q^{\mathrm{ehr}}, q^{\mathrm{img}},
q^{\mathrm{sens}}, \\
&q^{\mathrm{guide}}, q^{\mathrm{ref}}, q^{\mathrm{safety}}\}.
\end{aligned}
\label{eq:decomposition}
\end{equation}
For example, a referral query is decomposed into symptom extraction, longitudinal risk-history analysis, image abnormality interpretation, sensor-biomarker estimation, guideline matching, red-flag verification, and referral feasibility analysis. This design directly addresses long-context degradation, where relevant evidence may be hidden in the middle of long inputs or distributed across several data sources~\cite{b4,b5}.

Each modality-specific subtask is handled by a specialized clinical agent:
\begin{equation}
\begin{aligned}
z_p^m = F_m\left(X_p^m, q^m, \rho(q^m,\E_p)\right), \\
m \in \{\mathrm{text},\mathrm{ehr},\mathrm{img},
\mathrm{sens},\mathrm{guide}\}.
\end{aligned}
\label{eq:modality_agent}
\end{equation}
Here, the text agent extracts symptoms and clinical entities, the EHR agent models longitudinal coded events, the image agent interprets medical or smartphone-acquired images, the sensor agent computes digital biomarkers, and the guideline agent retrieves and verifies evidence-based recommendations.

\subsection{Clinical Evidence Graph Memory}

To ensure traceability and reduce unsupported generation, MedRLM builds a Clinical Evidence Graph Memory
\begin{equation}
\M_p = (\mathcal{V}_p, \mathcal{E}^{g}_p),
\label{eq:graph_memory}
\end{equation}
where nodes $v_i \in \mathcal{V}_p$ represent patient observations, clinical entities, abnormal image findings, sensor biomarkers, guideline statements, or referral criteria. Edges $e_{ij} \in \mathcal{E}^{g}_p$ represent temporal, semantic, causal, or guideline-based relationships.

Each evidence node is represented as a clinical triple
\begin{equation}
\tau_i = (o_i, s_i, \delta_i),
\label{eq:triple}
\end{equation}
where $o_i$ is the patient-specific observation, $s_i$ is the supporting source or guideline statement, and $\delta_i$ is the standardized clinical definition. The evidence relevance score is
\begin{equation}
\begin{aligned}
S(e_i|q,p) =
&\lambda_1\,\mathrm{sim}(h_q,h_{e_i})
+ \lambda_2\,\mathrm{rel}(e_i,p) \\
&+ \lambda_3\,\mathrm{cred}(e_i)
- \lambda_4\,\mathrm{age}(e_i) \\
&- \lambda_5\,\mathrm{conflict}(e_i),
\end{aligned}
\label{eq:evidence_score}
\end{equation}
where $\mathrm{sim}(\cdot)$ measures semantic similarity, $\mathrm{rel}(\cdot)$ measures patient-specific relevance, $\mathrm{cred}(\cdot)$ measures source credibility, $\mathrm{age}(\cdot)$ penalizes outdated evidence, and $\mathrm{conflict}(\cdot)$ penalizes contradiction.

The retrieved evidence representation is obtained using normalized evidence weights:
\begin{equation}
\alpha_i =
\frac{\exp(S(e_i|q,p))}
{\sum_{j=1}^{N}\exp(S(e_j|q,p))},
\label{eq:alpha}
\end{equation}
\begin{equation}
z^{\mathrm{rag}}_p =
\sum_{i=1}^{N}\alpha_i h_{e_i}.
\label{eq:rag_representation}
\end{equation}
This graph memory extends medical RAG by linking patient-specific multimodal evidence with guideline statements and referral rules, rather than retrieving isolated passages only~\cite{b6,b12,b13,b14}.

\subsection{Sensor-Guided Recursive Screening}

A key novelty of MedRLM is sensor-guided recursive triggering. Given a physiological or behavioral sensor stream
\begin{equation}
X^{\mathrm{sens}}_p = \{x_1,x_2,\ldots,x_T\},
\label{eq:sensor_stream}
\end{equation}
a temporal encoder extracts window-level digital biomarkers:
\begin{equation}
b_t = G_{\theta}\left(x_{t-w:t}\right),
\label{eq:biomarker}
\end{equation}
where $w$ is the temporal window and $b_t$ is the biomarker embedding at time $t$. Patient-specific abnormality is measured using a baseline-adjusted distance:
\begin{equation}
a_t =
(b_t-\mu_p)^{T}\Sigma_p^{-1}(b_t-\mu_p),
\label{eq:abnormality}
\end{equation}
where $\mu_p$ and $\Sigma_p$ denote the baseline distribution of patient $p$.

If $a_t$ exceeds a predefined threshold, the sensor signal activates deeper recursive reasoning:
\begin{equation}
\begin{aligned}
q^{\mathrm{sens}}_t =
\mathrm{Trigger}(a_t,\G,\R),
\quad \mathrm{if}\quad a_t > \epsilon .
\end{aligned}
\label{eq:sensor_trigger}
\end{equation}
Thus, an abnormal sensor pattern is not treated as a standalone classification result. It becomes a reasoning trigger that asks MedRLM to re-check symptoms, longitudinal history, image findings, guideline criteria, and referral rules.

\subsection{Multimodal Risk Estimation}

After recursive processing, MedRLM obtains modality-level representations and forms a fused patient state:
\begin{equation}
\begin{aligned}
Z_p = [&z^{\mathrm{text}}_p \Vert z^{\mathrm{ehr}}_p
\Vert z^{\mathrm{img}}_p \\
&\Vert z^{\mathrm{sens}}_p
\Vert z^{\mathrm{rag}}_p],
\end{aligned}
\label{eq:fusion}
\end{equation}
where $\Vert$ denotes concatenation. The risk score is estimated as
\begin{equation}
r_p = \sigma(W_r Z_p + b_r),
\label{eq:risk_binary}
\end{equation}
where $\sigma(\cdot)$ is the sigmoid function. For multi-class screening, the probability of class $c$ is
\begin{equation}
P(y=c|p) =
\frac{\exp(W_cZ_p+b_c)}
{\sum_{c'=1}^{C}\exp(W_{c'}Z_p+b_{c'})}.
\label{eq:softmax}
\end{equation}
The final clinical interpretation is selected as
\begin{equation}
\hat{y}_p = \arg\max_{c} P(y=c|p).
\label{eq:yhat}
\end{equation}

\subsection{Uncertainty-Gated Recursive Refinement}

MedRLM uses uncertainty to decide whether to answer, retrieve more evidence, recursively refine the reasoning path, or route the case to human review. The uncertainty score is defined as
\begin{equation}
\begin{aligned}
U_p =
&\beta_1\left(1-\max_c P(y=c|p)\right) \\
&+ \beta_2\,\mathrm{Var}\left(\{r_p^{(l)}\}_{l=1}^{L}\right) \\
&+ \beta_3(1-\bar{c}_p)
+ \beta_4\Gamma_p,
\end{aligned}
\label{eq:uncertainty}
\end{equation}
where the first term captures predictive uncertainty, the second term captures self-consistency variance across $L$ recursive trajectories, $\bar{c}_p$ is the average verbalized confidence, and $\Gamma_p$ is the evidence-conflict score. If $U_p$ exceeds a safety threshold $\delta$, MedRLM performs recursive refinement:
\begin{equation}
(q',\E'_p) =
\mathrm{Refine}(q,\E_p,\A_p,U_p),
\label{eq:refine}
\end{equation}
or routes the case to clinician review. This mechanism follows the insight that recursion alone is not sufficient for difficult semantic reasoning, and should be guided by uncertainty and self-reflection~\cite{b2,b3}.

\subsection{Community-to-Tertiary Referral Optimization}

The referral module models the care pathway as a constrained decision problem. The decision space is
\begin{equation}
\D = \{d_1,d_2,d_3,d_4\},
\label{eq:decision_space}
\end{equation}
where $d_1$ denotes self-care education, $d_2$ denotes follow-up at primary care, $d_3$ denotes specialist teleconsultation, and $d_4$ denotes tertiary-care referral.

The optimal referral decision is selected by maximizing expected clinical utility:
\begin{equation}
\begin{aligned}
d_p^{*} = \arg\max_{d \in \D}
\Big[&B(d,r_p,U_p,\mathcal{F}_p) \\
&- C(d,\mathcal{H})
- \eta D(d,r_p)\Big],
\end{aligned}
\label{eq:referral_utility}
\end{equation}
where $B(\cdot)$ is the expected clinical benefit, $C(\cdot)$ is the resource or system burden, $\mathcal{F}_p$ denotes red-flag findings, and $D(\cdot)$ penalizes delayed referral under high-risk conditions. This makes the framework suitable for community-to-tertiary referral support because it explicitly considers both patient risk and real-world healthcare constraints.

\subsection{Training Objective}

MedRLM can be optimized using supervised labels, weak evidence labels, retrieval relevance annotations, and clinician-preference feedback. The total objective is
\begin{equation}
\begin{aligned}
\mathcal{L} =
&\mathcal{L}_{\mathrm{cls}}
+ \lambda_{\mathrm{ref}}\mathcal{L}_{\mathrm{ref}}
+ \lambda_{\mathrm{rag}}\mathcal{L}_{\mathrm{rag}} \\
&+ \lambda_{\mathrm{align}}\mathcal{L}_{\mathrm{align}}
+ \lambda_{\mathrm{unc}}\mathcal{L}_{\mathrm{unc}}
+ \lambda_{\mathrm{safety}}\mathcal{L}_{\mathrm{safety}}.
\end{aligned}
\label{eq:loss_total}
\end{equation}
The classification and referral losses are
\begin{equation}
\mathcal{L}_{\mathrm{cls}} =
-\sum_{c=1}^{C} y_c\log P(y=c|p),
\label{eq:loss_cls}
\end{equation}
\begin{equation}
\mathcal{L}_{\mathrm{ref}} =
-\sum_{d\in\D} y_d^{\mathrm{ref}}\log P(d|p).
\label{eq:loss_ref}
\end{equation}
The retrieval ranking loss is
\begin{equation}
\begin{aligned}
\mathcal{L}_{\mathrm{rag}} =
-\log
\frac{\exp(S(e^{+}|q,p))}
{\exp(S(e^{+}|q,p))
+ \sum_{e^{-}}\exp(S(e^{-}|q,p))},
\end{aligned}
\label{eq:loss_rag}
\end{equation}
where $e^{+}$ denotes clinically relevant evidence and $e^{-}$ denotes irrelevant or weakly supported evidence.

The modality-alignment loss is formulated as
\begin{equation}
\begin{aligned}
\mathcal{L}_{\mathrm{align}} =
-\log
\frac{\exp(\mathrm{sim}(z_p^m,z_p^n)/\tau)}
{\sum_{j}\exp(\mathrm{sim}(z_p^m,z_j^n)/\tau)},
\end{aligned}
\label{eq:loss_align}
\end{equation}
where $z_p^m$ and $z_p^n$ are paired representations from two modalities, and $\tau$ is a temperature parameter. The safety loss penalizes recommendations that are unsupported by evidence or inconsistent with referral rules:
\begin{equation}
\mathcal{L}_{\mathrm{safety}} =
\mathbb{I}[d_p \notin \R(q,p)]
+ \xi\,\mathbb{I}[\A_p=\emptyset].
\label{eq:loss_safety}
\end{equation}

\subsection{Inference Algorithm}

Algorithm~\ref{alg:medrlm} summarizes the recursive inference process. The algorithm begins by initializing the external clinical environment, decomposes the input query into modality-specific subtasks, retrieves evidence, invokes specialized agents, recursively refines uncertain branches, builds a graph memory, estimates risk, and selects a referral decision using utility optimization.

\begin{algorithm}[t]
\caption{MedRLM Recursive Clinical Inference}
\label{alg:medrlm}
\begin{algorithmic}[1]
\Require Clinical query $q$, patient environment $\E_p$, context threshold $K$, safety threshold $\delta$
\Ensure Prediction $\hat{y}_p$, risk $r_p$, referral decision $d_p$, audit trail $\A_p$
\State Initialize external clinical environment $\E_p$
\State Decompose $q$ into subtasks $\mathcal{Q}_p$
\For{each subquery $q_j \in \mathcal{Q}_p$}
    \State Retrieve evidence $\rho(q_j,\E_p)$
    \If{$\kappa(q_j,\E_p) > K$}
        \State Split $q_j$ into smaller clinical subtasks
        \State Invoke MedRLM recursively on each subtask
    \Else
        \State Invoke the corresponding modality agent $F_m$
    \EndIf
\EndFor
\State Construct Clinical Evidence Graph Memory $\M_p$
\State Estimate risk score $r_p$ and evidence weights $\alpha_i$
\State Compute uncertainty $U_p$
\If{$U_p > \delta$}
    \State Perform recursive refinement or route to clinician review
\EndIf
\State Select $d_p^{*}$ using Eq.~\eqref{eq:referral_utility}
\State \Return $(\hat{y}_p,r_p,d_p^{*},\A_p,U_p)$
\end{algorithmic}
\end{algorithm}

\subsection{Methodological Novelty}

The methodological novelty of MedRLM is fourfold. First, it extends recursive language modeling from generic long-context reasoning to clinically grounded multimodal decision support. Second, it introduces a Clinical Evidence Graph Memory that links patient-specific observations, standardized clinical definitions, retrieved medical evidence, and referral criteria. Third, it proposes sensor-guided recursive triggering, where abnormal physiological or behavioral signals initiate deeper reasoning over symptoms, patient history, image findings, and guidelines. Fourth, it combines uncertainty-gated recursion with referral-utility optimization, allowing the system to decide when to answer, when to retrieve more evidence, when to defer to clinicians, and when to escalate from community care to tertiary care.

Unlike conventional medical LLM, medical RAG, or multimodal AI systems, MedRLM treats the patient case as a structured external environment and treats reasoning as a recursive clinical workflow. Therefore, the framework is scalable to long patient histories, robust to distributed evidence, compatible with multimodal health data, and auditable for safety-critical screening and referral support.

\section{Experiments and Results}

\subsection{Real Non-Synthetic Dataset Coverage}

\textbf{Can \MedRLM be evaluated on real clinical data without relying on synthetic patient cases?}
To answer this question, we selected datasets that are either public or available through credentialed research access, and that collectively cover the core inputs required by \MedRLM: long-context EHR, radiology images and reports, physiologic time series, ECG waveforms, multi-center ICU records, and referral-relevant risk outcomes. Table~\ref{tab:_real_dataset_inventory} summarizes the real datasets used for the  benchmark design. All sample sizes are taken from official dataset documentation or the corresponding dataset papers; no synthetic records are introduced.

\begin{table*}[t]
\centering
\caption{These datasets provide a non-synthetic evaluation basis for MedRLM across long-context EHR reasoning, image-report grounding, sensor-guided screening, and referral-relevant triage.}
\label{tab:_real_dataset_inventory}
\footnotesize
\begin{tabular}{p{0.12\textwidth}p{0.15\textwidth}p{0.18\textwidth}p{0.12\textwidth}p{0.23\textwidth}}
\toprule
\textbf{Dataset} & \textbf{Source} & \textbf{Real scale used in } & \textbf{Modality} & \textbf{Role in MedRLM evaluation} \\
\midrule
MIMIC-IV v3.1 & Beth Israel Deaconess Medical Center, PhysioNet & 364,627 individuals, 546,028 hospitalizations, and 94,458 ICU stays reported in the dataset documentation & EHR, ICU, emergency care & Long-context patient-history reasoning, diagnosis/procedure/lab retrieval, and risk-aware triage \cite{mimiciv2024,mimiciv2023}. \\
\addlinespace
MIMIC-CXR-JPG v2.1.0 & Beth Israel Deaconess Medical Center, PhysioNet & 377,110 chest radiographs associated with 227,827 imaging studies & Chest X-ray, radiology reports & Multimodal evidence grounding between image findings, reports, and patient context \cite{mimiccxrjpg2024,mimiccxr2019}. \\
\addlinespace
CheXpert & Stanford Hospital & 224,316 chest radiographs from 65,240 patients with uncertainty-aware labels & Chest X-ray, reports & External radiology benchmark for uncertainty labels and abnormality detection \cite{chexpert2019}. \\
\addlinespace
eICU-CRD v2.0 & Philips eICU program, PhysioNet & Over 200,000 ICU admissions from over 139,000 unique patients across 335 units and 208 US hospitals & Multi-center ICU EHR & External validation for ICU risk, tele-ICU workflow, and referral-proxy decisions \cite{eicu2019,eicu2018}. \\
\addlinespace
PTB-XL v1.0.3 & Physikalisch-Technische Bundesanstalt, PhysioNet & 21,799 clinical 12-lead ECGs from 18,869 patients with cardiologist annotations & ECG waveform & Sensor-guided recursive screening and abnormal cardiac-signal triggering \cite{ptbxl2022,ptbxl2020}. \\
\addlinespace
PhysioNet/CinC Challenge 2012 & PhysioNet/Computing in Cardiology & 12,000 adult ICU stays; set A provides 4,000 labeled training records with up to 42 variables in the first 48 hours & ICU vital-sign time series & Mortality-risk prediction, calibration, and clinically meaningful risk stratification \cite{challenge2012,physionet2000}. \\
\bottomrule
\end{tabular}
\end{table*}

The  result is positive: the selected benchmark covers every major evidence channel required by \MedRLM. MIMIC-IV and eICU-CRD support long-context clinical history and external ICU validation; MIMIC-CXR-JPG and CheXpert support image-report grounding; PTB-XL and the PhysioNet/CinC 2012 challenge support physiologic signal and time-series screening. The main remaining limitation is that direct community-to-tertiary referral labels are uncommon in public datasets. Therefore, referral experiments should use clinically defensible proxy outcomes such as ICU admission, in-hospital mortality, acute deterioration, readmission, specialist escalation, or tele-ICU intervention until a referral-labeled regional dataset is available.

\begin{figure}[t]
\centering
\begin{tikzpicture}[x=2.35cm,y=0.47cm]
\footnotesize
\draw[->,mSlate] (0,-0.15) -- (2.25,-0.15) node[right]{\scriptsize record scale};

\foreach \x/\lab in {0.1/$10^4$,1.1/$10^5$,2.1/$10^6$} {
  \draw[mSlate] (\x,-0.24) -- (\x,-0.06);
  \node[below,mSlate] at (\x,-0.27) {\scriptsize \lab};
}

\node[anchor=east] at (-0.05,6.0) {MIMIC-IV};
\draw[fill=mBlue,draw=none] (0,5.82) rectangle (1.84,6.18);
\node[anchor=west] at (1.88,6.0) {\scriptsize 546k hospitalizations};

\node[anchor=east] at (-0.05,5.1) {MIMIC-CXR};
\draw[fill=mCyan,draw=none] (0,4.92) rectangle (1.68,5.28);
\node[anchor=west] at (1.72,5.1) {\scriptsize 377k radiographs};

\node[anchor=east] at (-0.05,4.2) {CheXpert};
\draw[fill=mTeal,draw=none] (0,4.02) rectangle (1.45,4.38);
\node[anchor=west] at (1.49,4.2) {\scriptsize 224k radiographs};

\node[anchor=east] at (-0.05,3.3) {eICU};
\draw[fill=mAmber,draw=none] (0,3.12) rectangle (1.40,3.48);
\node[anchor=west] at (1.44,3.3) {\scriptsize $>$200k admissions};

\node[anchor=east] at (-0.05,2.4) {PTB-XL};
\draw[fill=mOrange,draw=none] (0,2.22) rectangle (0.44,2.58);
\node[anchor=west] at (0.48,2.4) {\scriptsize 21.8k ECGs};

\node[anchor=east] at (-0.05,1.5) {CinC 2012};
\draw[fill=mPurple,draw=none] (0,1.32) rectangle (0.18,1.68);
\node[anchor=west] at (0.22,1.5) {\scriptsize 12k ICU stays};

\node[
  anchor=west,
  text=mSlate,
  text width=7.2cm,
  align=left
] at (0,0.72)
{\scriptsize Bar length is proportional to $\log_{10}(N)-3.9$;\\
labels show the real dataset scale.};
\end{tikzpicture}
\caption{The chart visualizes the practical evaluation breadth available without synthetic cases.}
\label{fig:_dataset_scale}
\end{figure}
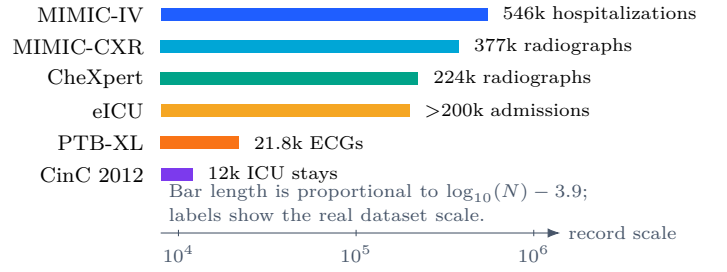

\subsection{Benchmark Tasks and Metrics}

Table~\ref{tab:_tasks_metrics} maps the real datasets to measurable evaluation tasks. The design intentionally separates three claims: (i) dataset availability and coverage, which is reported in this section; (ii) published benchmark anchors, which are cited from the original benchmark pages; and (iii) future MedRLM performance, which should only be reported after a full run of Algorithm~1 on the protected datasets.

\begin{table*}[t]
\centering
\caption{ benchmark tasks, labels, and metrics grounded in real datasets.}
\label{tab:_tasks_metrics}
\footnotesize
\begin{tabular}{p{0.16\textwidth}p{0.14\textwidth}p{0.18\textwidth}p{0.16\textwidth}p{0.16\textwidth}}
\toprule
\textbf{Evaluation task} & \textbf{Dataset(s)} & \textbf{Target label or output} & \textbf{Primary metric(s)} & \textbf{Clinical interpretation} \\
\midrule
Long-context EHR risk reasoning & MIMIC-IV, eICU-CRD & Mortality, ICU admission, readmission, length of stay, diagnosis/procedure events & AUROC, AUPRC, calibration error, decision-curve net benefit & Tests whether recursive retrieval improves risk estimation over compressed one-shot summaries. \\
\addlinespace
Image-report grounding & MIMIC-CXR-JPG, CheXpert & Radiology abnormality labels, uncertain findings, image-report consistency & Mean AUROC, macro-F1, unsupported-claim rate, evidence precision & Tests whether the evidence graph links image findings to report statements and clinical context. \\
\addlinespace
Sensor-guided screening & PTB-XL, PhysioNet/CinC 2012 & ECG diagnostic statements, abnormal vital-sign trajectories, in-hospital death & Macro-F1, AUROC, sensitivity at fixed specificity, under-triage rate & Tests whether abnormal signals trigger deeper recursive reasoning rather than isolated classification. \\
\addlinespace
Referral-proxy decision support & MIMIC-IV, eICU-CRD, PhysioNet/CinC 2012 & ICU escalation, mortality risk, deterioration proxy, tele-ICU intervention proxy & Referral utility, over-referral rate, under-referral rate, calibration within risk deciles & Evaluates whether referral recommendations balance patient safety and resource burden. \\
\addlinespace
Auditable evidence synthesis & Linked MIMIC-IV/MIMIC-CXR and retrieved medical evidence & Evidence triples, source attribution, conflict flags, final recommendation rationale & Evidence recall, citation precision, conflict-detection rate, clinician review rate & Tests whether MedRLM produces traceable reasoning instead of unsupported free-text answers. \\
\bottomrule
\end{tabular}
\end{table*}

\subsection{Published Benchmark Anchors}

Because the complete \MedRLM pipeline has not yet been executed on the credentialed datasets, Table~\ref{tab:published_anchors} reports published real-data anchors rather than invented MedRLM accuracy. These values define minimum comparison points for future experiments. On the PhysioNet/CinC 2012 mortality benchmark, the official challenge page reports that the best Event 1 score on the hidden validation set C was 0.5353, SAPS-I scored 0.3125, and a random predictor scored 0.1386. For calibrated risk estimation in Event 2, lower is better: the best published score was 17.88, SAPS-I scored 68.58, and a random predictor scored 10137.7 \cite{challenge2012}. These anchors are important because they make the results section falsifiable: a future MedRLM implementation should be compared against established real-data baselines, not against synthetic case studies.

\begin{table}[t]
\centering
\caption{Published real-data anchors from PhysioNet/CinC Challenge 2012. These are not claimed as MedRLM results; they are comparison points for future MedRLM experiments.}
\label{tab:published_anchors}
\footnotesize
\begin{tabular}{p{0.36\columnwidth}p{0.24\columnwidth}p{0.24\columnwidth}}
\toprule
\textbf{Benchmark anchor} & \textbf{Event 1 score} & \textbf{Event 2 score} \\
\midrule
Best published participant & 0.5353 & 17.88 \\
SAPS-I reference model & 0.3125 & 68.58 \\
Random predictor & 0.1386 & 10137.7 \\
\bottomrule
\end{tabular}
\end{table}

\end{document}